\documentclass{article}
\usepackage{spconf,amsmath,graphicx}
\usepackage{subfigure}
\usepackage{comment}
\usepackage[table]{xcolor}
\usepackage{multirow}
\usepackage{xcolor}

\def\eg{\textit{e.g.\,}}
\def\ie{\textit{i.e.\,}}
\def\etc{\textit{etc.\,}}

\newcommand{\rex}[1]{\textcolor{red}{#1}}
\usepackage{amssymb}
\usepackage{pifont}
\newcommand{\cmark}{\ding{51}}%
\newcommand{\xmark}{\ding{55}}%
\newcommand{\layer}[1]{{\small{\texttt{#1}}}}
\usepackage{caption}
\usepackage{hyperref}
\hypersetup{
    colorlinks=true,
    linkcolor=magenta,
    filecolor=blue,      
    urlcolor=magenta,
}
\usepackage[hang,flushmargin]{footmisc}
\title{\texttt{LineCounter}: Learning Handwritten Text Line Segmentation by Counting        }

\name{Deng Li$^{\star}$ \qquad Yue Wu$^{\dagger}$ \qquad Yicong Zhou$^{\star}$ \thanks{This work was done prior to Amazon involvement of the authors.}
\thanks{This work was funded by The Science and Technology Development Fund, Macau SAR (File no. 189/2017/A3), and by University of Macau (File no. MYRG2018-00136-FST).}}

\address{$^{\star}$ University of Macau, Department of Computer and Information Science, Macau, China\\
$^{\dagger}$ Amazon Alexa Natural Understanding, Manhattan Beach, CA, USA\\
\small{\texttt{mb85511@um.edu.com}, \,
\texttt{wuayue@amazon.com}, \,
\texttt{yicongzhou@um.edu.mo}}
}

\begin{document}

%
\maketitle
\begin{abstract}
Handwritten Text Line Segmentation (HTLS) is a low-level but important task for many higher-level document processing tasks like handwritten text recognition. It is often formulated in terms of semantic segmentation or object detection in deep learning. However, both formulations have serious shortcomings. The former requires heavy post-processing of splitting/merging adjacent segments, while the latter may fail on dense or curved texts. In this paper, we propose a novel \textit{Line Counting} formulation for HTLS -- that involves counting the number of text lines from the top at every pixel location. This formulation helps learn an end-to-end HTLS solution that directly predicts per-pixel line number for a given document image. Furthermore, we propose a deep neural network (DNN) model \layer{LineCounter} to perform HTLS through the \textit{Line Counting} formulation. Our extensive experiments on the three public datasets (ICDAR2013-HSC~\cite{2013icdar}, HIT-MW~\cite{su2007corpus}, and VML-AHTE~\cite{barakat2020unsupervised}) demonstrate that \layer{LineCounter} outperforms state-of-the-art HTLS approaches. Source code is available at~\url{https://github.com/Leedeng/Line-Counter}.
\end{abstract}
\begin{keywords}
handwritten text line segmentation, deep learning, counting, OCR, document analysis
\end{keywords}
\section{Introduction}
\label{sec:Introduction}
The Handwritten Text Line Segmentation (HTLS) problem is a classic problem in document analysis that serves as a prerequisite step for many higher-level document processing tasks like keyword spotting~\cite{giotis2017survey}, table analysis~\cite{Gilani2017table} and handwritten text recognition~\cite{plamondon2000online}. The HTLS problem is very challenging due to the possible large variations of text lines in a document, not to speak of those written by different authors and/or in different script-languages. Such variations include but not limited to line orientation, line/word spacing, font size, script language, writing style, image quality, \etc~\cite{likforman2007text,li2008script}.\par

\begin{figure}[!htb]
  \centering
  \scriptsize
  \setlength{\fboxsep}{0pt}
  \setlength\tabcolsep{.5pt}
  \def\lh{2.3cm}
  \def\lw{2.7cm}
  \begin{tabular}{@{}cccc@{}}
       \fbox{\includegraphics[trim={.15cm .15cm .15cm .15cm}, clip, height=\lh,width=\lw]{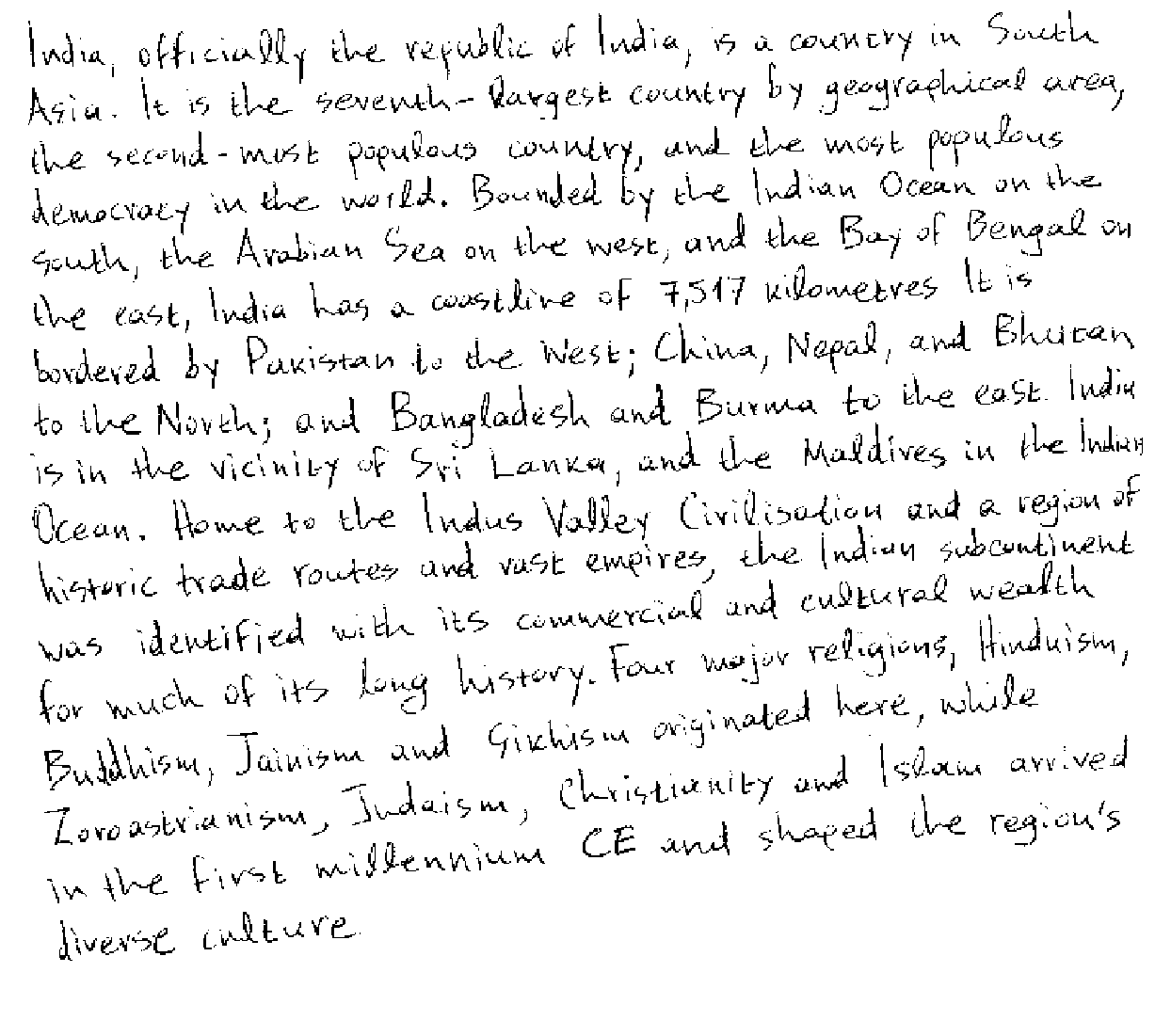}}&
       \fbox{\includegraphics[height=\lh,width=\lw]{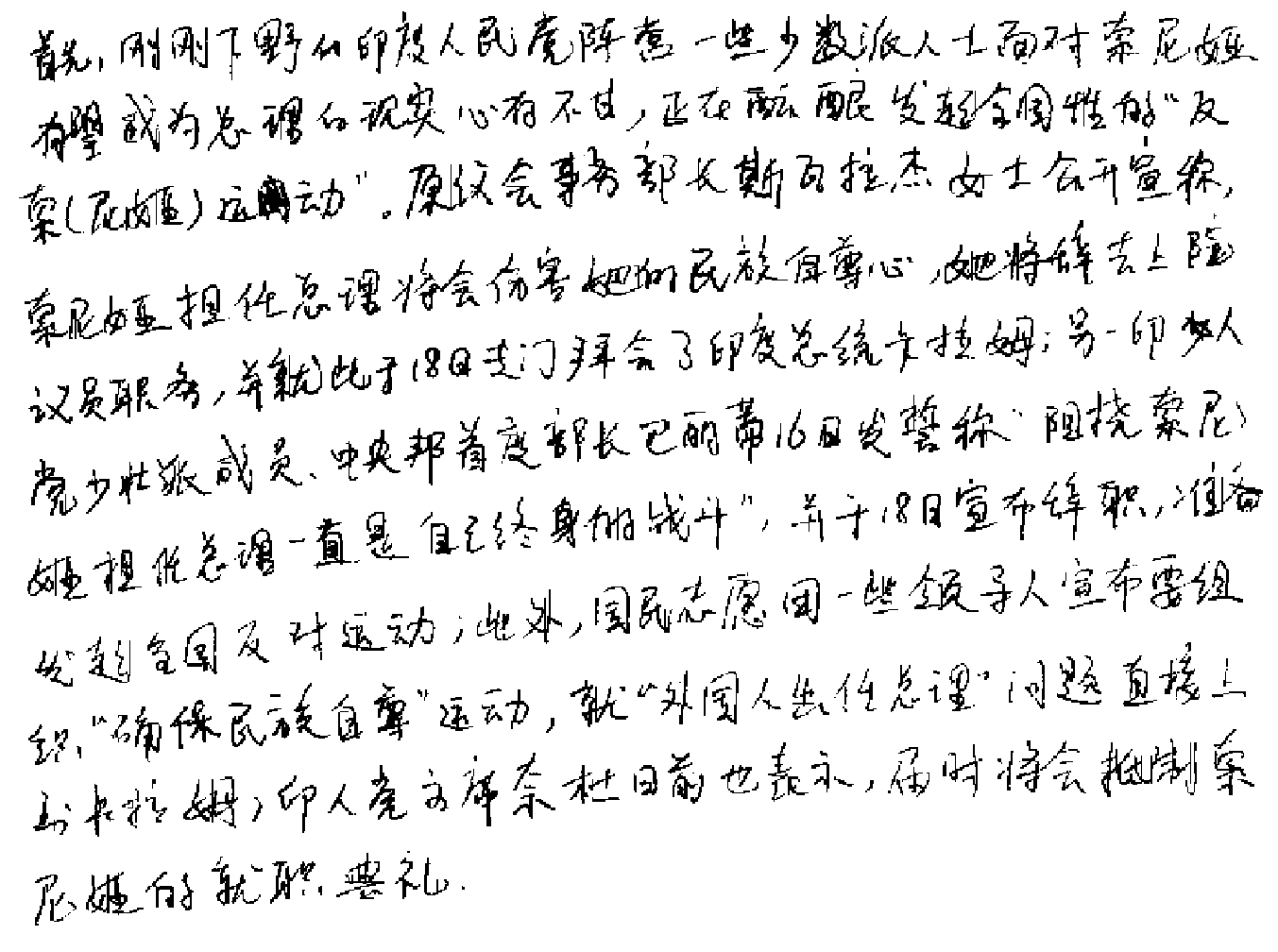}}&
       \fbox{\includegraphics[trim={5cm 12cm 1.5cm 5cm},clip,height=\lh,width=\lw]{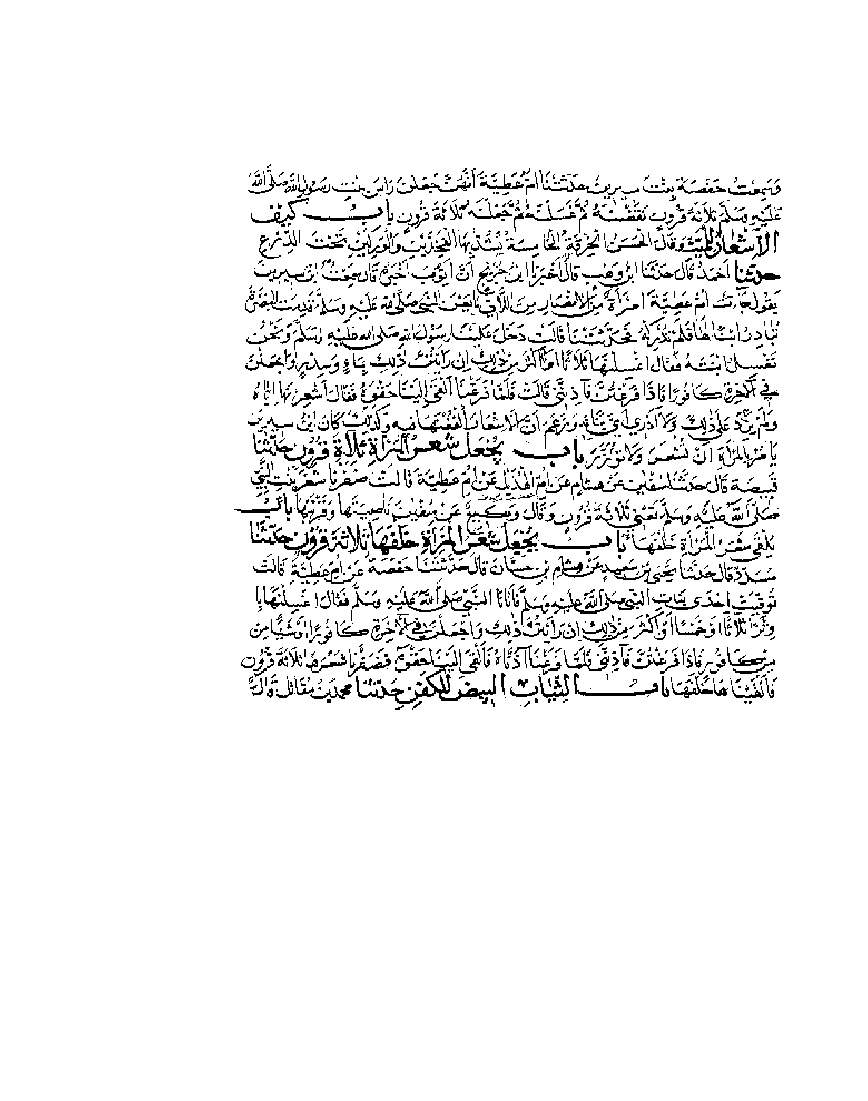}}&\\
       
       \fbox{\includegraphics[trim={.15cm .15cm .15cm .15cm}, clip, height=\lh,width=\lw]{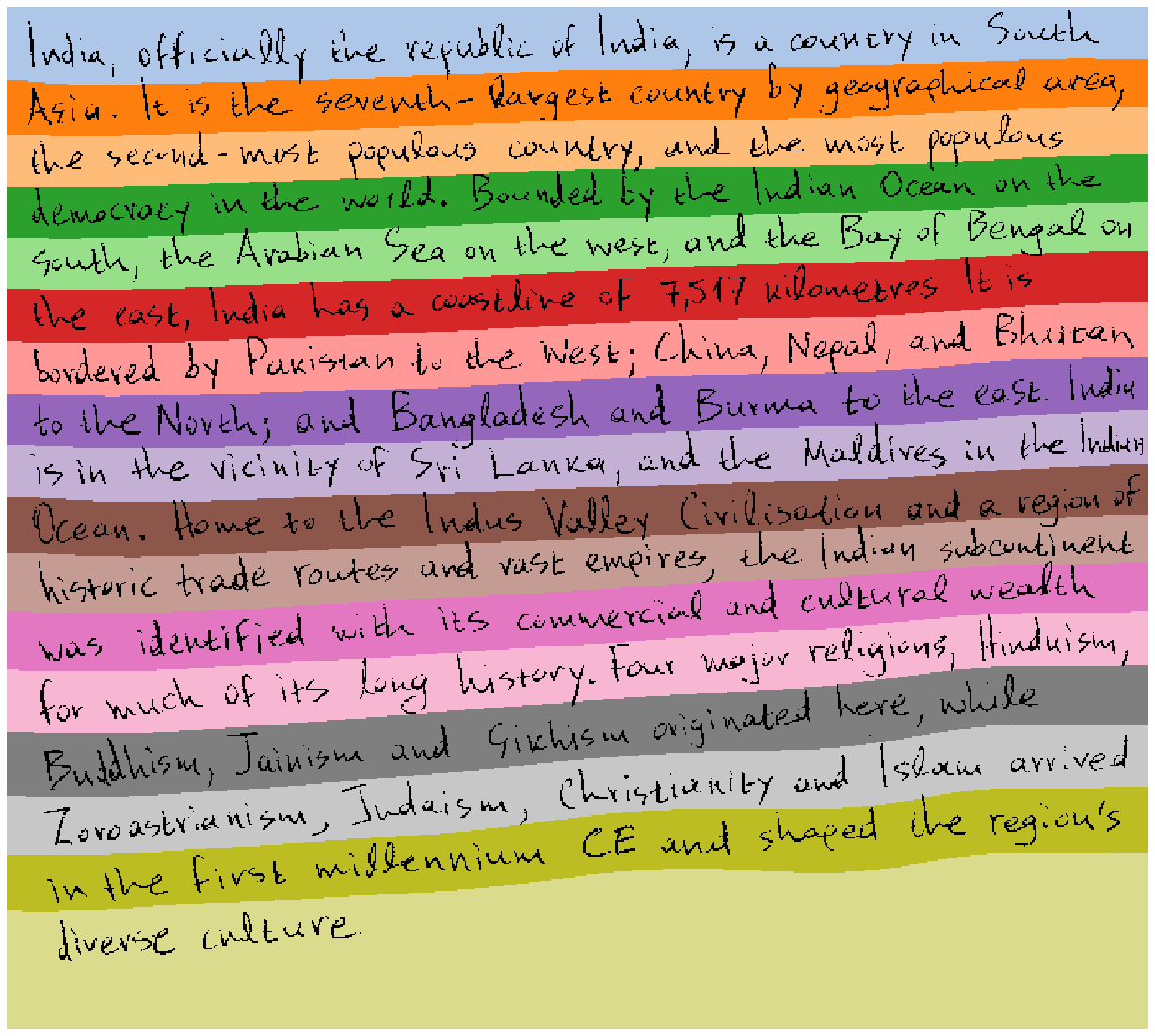}} & 
       \fbox{\includegraphics[height=\lh,width=\lw]{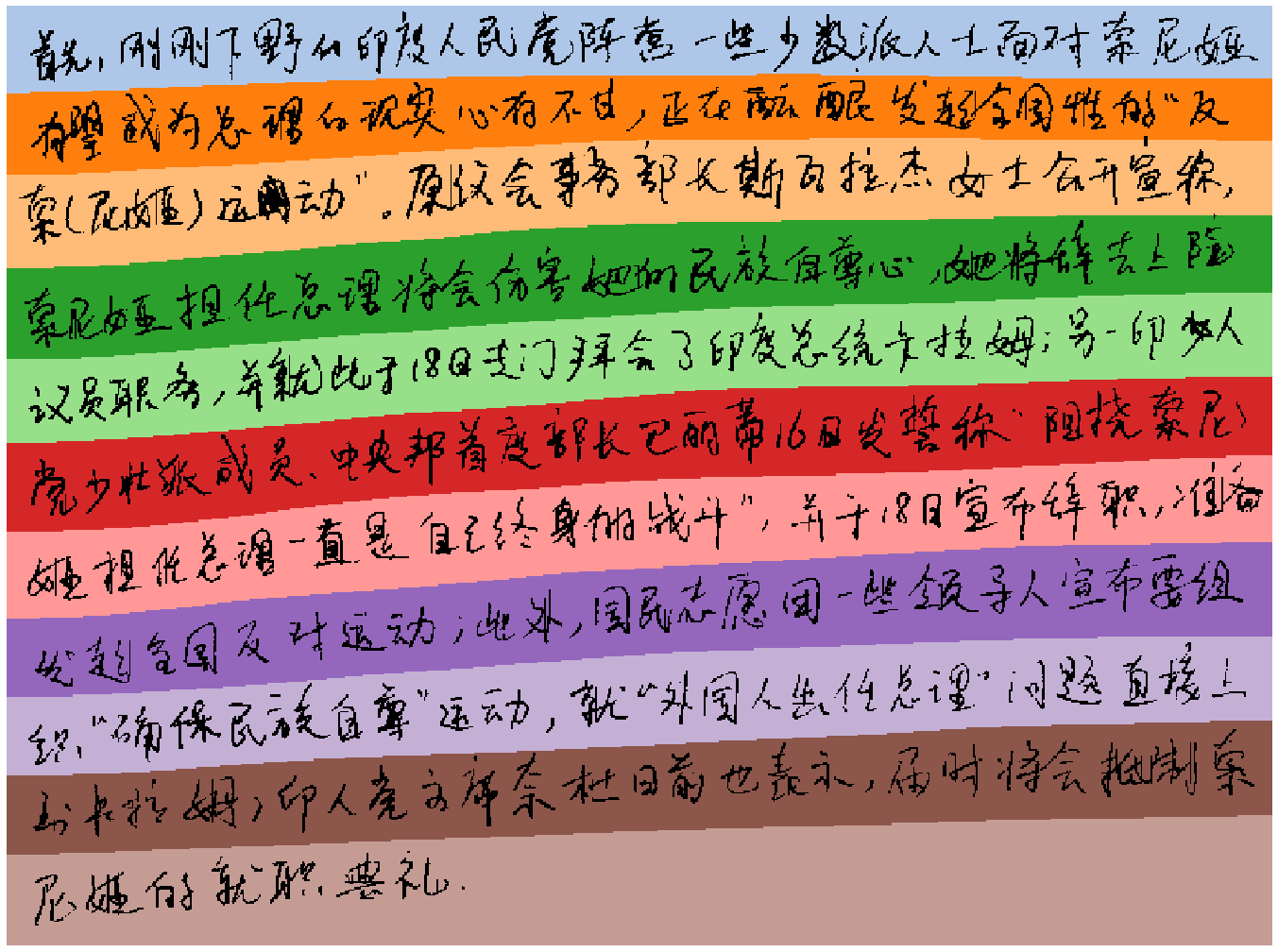}} &
       \fbox{\includegraphics[trim={5cm 12cm 1.5cm 5cm},clip,height=\lh,width=\lw]{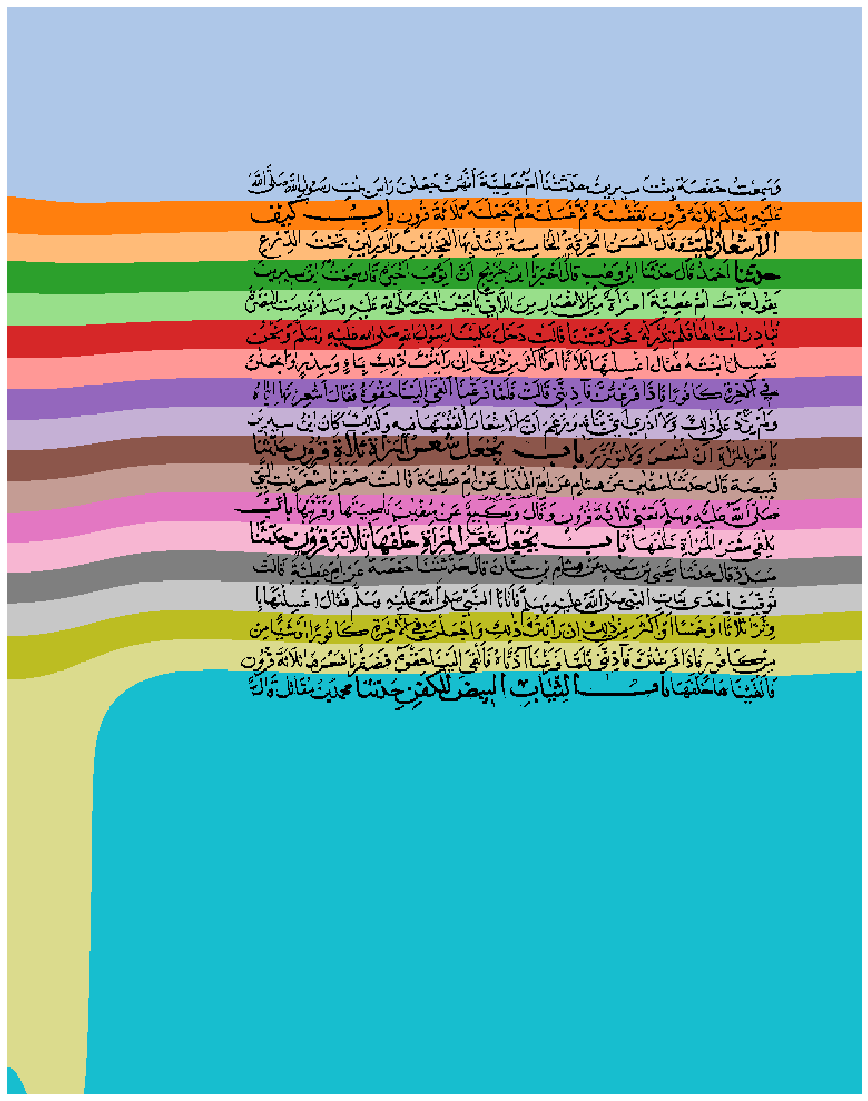}} &
       {\includegraphics[height=2.32cm]{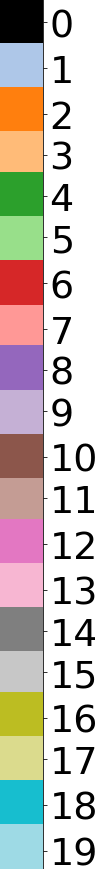}}\\
  \end{tabular}
  \caption{Sample \layer{LineCounter} input document images (upper) and output per-pixel line counting maps (lower).  Samples from left to right are from the ICDAR2013-HSC~\cite{2013icdar}, HIT-MW~\cite{su2007corpus}, and VML-AHTE dataset~\cite{barakat2020unsupervised}, respectively. 
  {Predicted line numbers are rounded to nearest integers and color-coded according to the rightmost color scheme. Texts are artificially superimposed on predictions for better visualization. Best viewed digitally in color and zoomed-in.}}\label{fig.result}
\end{figure}

Conventional solutions~\cite{wong1982document,ha1995document,wu2014markov} typically work under a constrained setup or require strong assumptions on one or more factors. For example, projection-based methods~\cite{ha1995document} analyze the 1D projection profile and solve the HTLS problem by using local minimum $y$-locations to separate consecutive lines. However, its success relies on whether a document image is rectified and whether the line length variation is small.
Smearing-based methods~\cite{wong1982document} use run-length analysis to connect text pixels belonging to the same line, but require the text line variations to be small within a page. For detailed reviews of the classic HTLS methods, please refer to \cite{likforman2007text}.

HTLS datasets are typically annotated in terms of line-maps ~\cite{2013icdar,barakat2020unsupervised,alberti2019labeling}, where each text pixel is annotated with a line number and all text pixels belonging to the same line share the same line number. Given an input document image $D$, the corresponding HTLS line-map $M$ is defined as follows.

\begin{equation}\label{eq.htls_gt}
M(x,y) = \begin{cases}
	k>0, & \textrm{if\,} (x, \!y) \textrm{\,is a text pixel in the\,} k\textrm{-th line}\\  
	0, &\text{if\,} (x, \!y) \textrm{\,is a non-text pixel in\,} D
		   \end{cases}
\end{equation}


It is the great success of the deep neural network (DNN) that inspires trainable end-to-end HTLS solutions. Common DNN-based HTLS problem formulations include: 1) semantic segmentation~\cite{alberti2019labeling,vo2017text,renton2018fully,barakat2018text},
2) object detection~\cite{mxnetHWR,sui2018novel}, and 3) sequence learning~\cite{moysset2015paragraph}. However, none of these approaches were originally designed for the HTLS task. Consequently, one has to fill the gap between the HTLS line map ground truth (GT) defined in Eq.~\eqref{eq.htls_gt} and the GT required for the adopted formulation. More precisely, one has to convert the HTLS line map GT to a desired GT before training and then convert prediction back to HTLS line map. Though such conversions seem to be trivial, both can be problematic. For example, one needs to apply a smearing kernel on an HTLS line map to obtain the required binary or ternary line blob mask for the semantic segmentation based HTLS ~\cite{vo2017text,renton2018fully}. For different smearing kernel sizes, one can get different line blob mask GTs. They are equivalent in the HTLS context but different for semantic segmentation, implying the GT conversion could be ambiguous. As another example, the object detection based HTLS problem formulation~\cite{sui2018novel,mxnetHWR} outputs line bounding boxes/masks, but a text pixel could belong to multiple predicted bounding boxes/masks. Thus one has to apply additional post-processing to decide its membership in the HTLS line map. Table~\ref{tab:related} summarizes recent DNN-based HTLS problem formulations.

In this paper, we make three main contributions: 1) we introduce the novel \textit{Line Counting} formulation for the HTLS task; 2) we propose the \layer{LineCounter} network that is designed specifically for the \textit{LineCouning} formulation; and 3) we re-invent the \layer{cumsum} activation to enforce monotonic outputs, which are beneficial for the \textit{LineCounting} learning. Consequently, we can train an end-to-end HTLS DNN using the line map directly and predict per-pixel line numbers to fulfill the HTLS task without annoying post-processing like splitting or voting. Sample results are shown in Fig.~\ref{fig.result}.

\begin{figure*}[bhtp]
    \centering
    \definecolor{Apricot}{RGB}{251,206,177}
    \definecolor{Goldenrod}{RGB}{253,243,177}
    \setlength{\fboxsep}{1pt}
    \includegraphics[trim=1cm 0 2cm 0, clip, width=\linewidth]{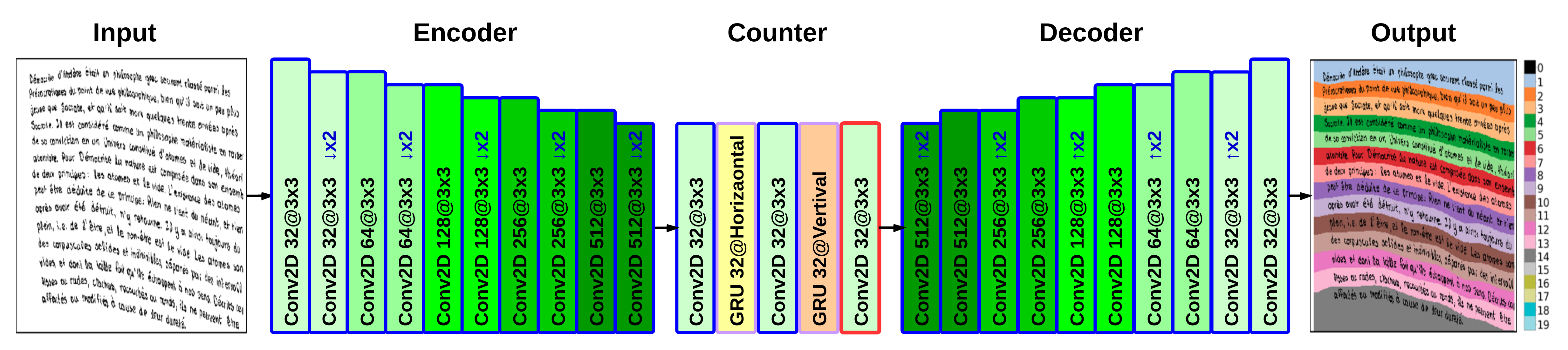}
    \caption{The proposed \texttt{LineCounter} network architecture. Each {\colorbox{green}{\layer{Conv2D}}} block with \textcolor{blue}{blue} border represents a module of \layer{Conv2D} + \layer{BatchNorm} + \texttt{ReLU}, convolution parameters are of format {\scriptsize{\$\{\layer{filters}\}@\$\{\layer{ksize}\}$\times$\$\{\layer{ksize}\}}}, and a darker block indicates more filters. Arrows denote down-sampling ($\downarrow$) or up-sampling ($\uparrow$).  \colorbox{Apricot}{\layer{GRU}} and \colorbox{Goldenrod}{\layer{GRU}} apply the bidirectional recurrent analysis by treating the time axis as the column and row axes, respectively. They both use \layer{hard\_sigmoid} recurrent activation, and \layer{tanh} output activation. \colorbox{green}{\layer{Conv2D}} block with \textcolor{red}{red} border is a convolutional layer with novel \layer{cumsum} activation introduced in Sec.~\ref{sec.counter}. 
    The output line number map is rounded and color coded according to the rightmost coloring scheme, and texts on the map are artificially superimposed for better visualization.
    }

    \label{fig.structure}
\end{figure*}

\begin{table}[!t]
\centering
\scriptsize
\caption{Existing deep learning based HTLS problem formulations.}\label{tab:related}
\setlength\tabcolsep{2pt}
\begin{tabular}{@{}r|r|r|r|r|r}
     \hline\hline
     {\textbf{Related}}& \textbf{Problem} & \textbf{DNN } &  {\textbf{Loss}}&\textbf{Unique}&{\textbf{Postproc.}}\\
     {\textbf{Works}}& \textbf{Formulation} & \textbf{Target Type} &  {\textbf{Type}}&\textbf{Target?}&\textbf{Free?}\\
     \hline
     ~\cite{vo2017text,renton2018fully,barakat2018text}& Semantic Seg.& Semantic Mask & Cls. & \xmark & \xmark \\
     ~\cite{mxnetHWR,sui2018novel}& Object Det. & BBox+Mask & Cls.+Reg.  &\xmark &  \cmark \\
     ~\cite{moysset2015paragraph}& Seq. Learning & Symbol Sequence  & CTC &  \cmark& \xmark\\
     ~\cite{wigington2018start}& Start-and-Follow & Boundary Coord. & Reg. &\xmark &  \cmark\ \\ \hline
     Ours & Line Counting & HTLS Line-map & Reg. & \cmark & \cmark \\
     \hline\hline
\end{tabular}
\end{table}

\section{The Line-Counting Formulation}
\label{sec:Proposed}
\subsection{Motivation}
As shown in Table~\ref{tab:related}, existing DNN based HTLS problem formulations may suffer two issues: 1) in training, the used network target derived from the HTLS line map might not be unique; and 2) during inference, the network's prediction may require heavy postprocessing to convert to HTLS line map. Both are related to the fact that an existing formulation is originally not designed for the HTLS task. Hence, in order to avoid these issues, a new problem formulation should be designed specifically for the HTLS task.

\subsection{The Line Counting Formulation for HTLS}
A straightforward solution is to formulate the HTLS task as a per-pixel regression task, where the input is the document image $D$, and the target is the HTLS line-map $M$ defined in Eq.~\eqref{eq.htls_gt}. Consequently, a DNN $f_{\textrm{DNN}}$ can be trained by supplying pairwise input $D$ and target $M$ as shown below.
\begin{equation}
    M = f_{\textrm{DNN}}(D)
\end{equation}
Unfortunately, $f_{\textrm{DNN}}$ does not work well in practice, possibly due to the absence of semantic information about those non-text sub-classes (\eg \textit{background}, \textit{noise}, \textit{space in between two lines/words}, \etc) Intuitively, one need such information to effectively differentiate adjacent text lines. Although it is still possible to learn such semantic information, existing solutions like~\cite{wu2017self} require to formulate the HTLS task again as a semantic segmentation task.

In this paper, we formulate the HTLS task as the \textit{Line Counting} problem. Formally, given a document image $D$ of size $H\times W$, we want to train a $g_{\textrm{DNN}}$ to predict a line counting map $C$ of the same size, where $C(x,y)$ indicates 
\textbf{the line number at location $(x,y)$ when counting from the top},
\begin{equation}
    C = g_{\textrm{DNN}}(D)
\end{equation}
and ideally we want predicted line numbers to be monotonically increasing along $y$-axis as shown in Eq.~\eqref{eq.monotonic}.
\begin{equation}\label{eq.monotonic}
    C(x,y) \leq C(x,y+1), \forall (x,y) \in \mathbb{R}^{H\times W}
\end{equation}
It is worthy noting that we can't fully derive the line counting map $C$ from the HTLS line-map $M$, but we have GTs for locations in the text pixel set $\mathbb{T}$ as shown in Eq.~\eqref{eq.CM}.
\begin{equation}\label{eq.CM}
   C^{GT}(x,y)=M(x,y), \textrm{\,if\,} (x, \!y) \in \mathbb{T}
\end{equation}
\noindent However, this doesn't prevent us to train  $g_{\textrm{DNN}}$ since the train loss can still be computed from text pixel locations as shown in Eq.~\eqref{eq.loss}, and the semantics about non-text classes can be implicitly learned via the constraint~\eqref{eq.monotonic}.
\begin{equation}\label{eq.loss}
\textrm{Loss} = \textstyle\sum_{(x,y) \in \mathbb{T}} | C(x,y) - C^{GT}(x,y) | /\|\mathbb{T}\|
\end{equation}

\subsection{The \texttt{LineCounter} Solution for HTLS}
To validate the \textit{Line Counting} formulation, we need to first design a compatible network. Unfortunately, most existing DNNs are not even close. We therefore propose our own DNN solution named \layer{LineCounter}. 

As shown in Fig.~\ref{fig.structure}, \layer{LineCounter} is a variant of the classic Encoder-Decoder network~\cite{cho2014learning} with three main modules, namely, \layer{Encoder}, \layer{Counter}, and \layer{Decoder}. It takes a image $D$ as input, and extracts convolutional feature using \layer{Encoder}. Next, it propagates line information along the horizontal and vertical directions in \layer{Counter}. Finally, it predicts the line number for each pixel location via \layer{Decoder}. 

We borrow the \layer{Encoder} design from DnCNN~\cite{zhang2017beyond}, and design \layer{Decoder} as the mirror of \layer{Encoder} (see Fig.~\ref{fig.structure} for layerwise settings). However, it is clear that just having \layer{Encoder} and \layer{Decoder} can't achieve the line counting goal, simply because it predicts identical line numbers for two identical patches regardless their relative locations. This is why \layer{Counter} is a must-have module for counting. 

As shown in Fig~\ref{fig.structure}, \layer{Counter} is a sub-network of five blocks: \layer{Conv2D}, \layer{GRU}, \layer{Conv2D}, \layer{GRU}, and \layer{Conv2D}. Three \layer{Conv2D} blocks (left to right) transform features between \layer{Encoder} and \layer{Counter}, horizontal and vertical \layer{GRU} blocks, and \layer{Counter} and \layer{Decoder}, respectively. 

In order to enable the recurrent sequence learning for a 3D (\ie height, width, and depth) image feature, we treat one spatial dimension as the time dimension and the other as the batch dimension. This allows \layer{GRU} to propagate global line information along the vertical (height as time) or horizontal (width as time) direction via recurrent sequence learning. The exchange of horizontal line information helps attain similar line numbers for pixels belonging to the same line. The exchange of vertical line information helps the network know the number of text lines have been seen at any point. Furthermore, it helps learn what white spaces are essential to line counting and what are not. In practice, we find that using the bi-directional setting and placing horizontal analysis first gives better performance (see Sec.~\ref{sec.counter}).  

Regarding the monotone property \eqref{eq.monotonic}, we find a surprisingly simple solution -- re-purposing the \layer{cumsum} function for activation. More precisely, \layer{cumsum} is defined as the cumulative sum for feature map $F$ along $y$-axis as shown in ~\eqref{eq.cumsum}. 
\begin{equation}\label{eq.cumsum}
    F'(x,y) = \layer{cumsum}(F)(x,y) = \textstyle\sum_{i=0}^{y} F(x,i)
\end{equation}
When $F$ is non-negative, \ie $F(x,y)\geq 0$, we have \eqref{eq.cumsum2}.
\begin{equation}\label{eq.cumsum2}
F'(x,y) \leq F'(x,y+1), \forall (x,y) \in \mathbb{R}^{H\times W}
\end{equation}
In practice, we find that it is better to enforce the feature monotone in the last \layer{Conv2D} of \layer{Counter} and use \layer{hard sigmoid} to ensure non-negative $F$ (see Sec.~\ref{sec:monotone}).




\section{Experimental Results}
\label{sec:Experiment}
\subsection{Experiment Settings}\label{sec.ablation}
\noindent\textbf{Data:} In total, three public dataset are used in our study, namely, ICDAR-HCS2013~\cite{2013icdar} (Latin and Bangla, 200/150), HIT-MW~\cite{su2007corpus} (Chinese, 747/106) and VML-AHTE~\cite{barakat2020unsupervised}  (Arabic, 20/10), where the content in braces indicates a dataset's script-language and training/testing sizes. We use the random perspective and thin plate spline transforms for data augmentation. See more details in our code repository. 

\noindent\textbf{Metrics:} we follow the classic HTLS evaluation protocols~\cite{2013icdar} and report performance in \textit{Detection Rate} (DR), \textit{Recognition Accuracy} (RA) and \textit{F-Measure} (FM). The one-to-one matching threshold is set to 0.9. Model speed is measured in terms of frame-per-second (FPS) on a \layer{Nvidia 2080Ti} GPU.

\noindent\textbf{\layer{LineCounter}:} we implement it in \layer{TensorFlow}. Training input size is set to 1088$\times$768 (A4 paper aspect ratio). We apply the aspect ratio preserving resize to a sample and pad white spaces if necessary. We set the batch size to 4, and use the \layer{Adam} optimizer with the initial learning rate 1$e$-4. 
This learning rate is halved if no improvement is observed for 20 epochs. 
For inference, we always resize an input as in training and round predicted line numbers to nearest integers.

\subsection{Ablation Study on \layer{Counter} Topology}\label{sec.counter}
As one can see, \layer{Counter} must propagate line number information, but it is unclear 1) which direction comes first; and 2) uni- or bi-directional recurrent sequence learning. To better answer these questions, we study the two \layer{GRU} blocks in \layer{Counter} while making the rest of network architecture unchanged -- all models in this study use the same \layer{Encoder} and \layer{Decoder} structures as shown in Fig.~\ref{fig.structure}, and same \layer{Conv2D} blocks (namely, \layer{Conv2D} +\layer{BatchNorm}+\layer{ReLU}) for \layer{Counter}. 


\begin{table}[!h]
\setlength{\tabcolsep}{8pt}
\centering
\caption{Ablation Study on Counter Topology}\label{tab:recurrent_layer}
\scriptsize 
\begin{tabular}{r@{}c| r@{}c|c}
\hline\hline
\multicolumn{2}{c}{\textbf{1$^{\tiny\textrm{st}}$ \layer{GRU}}} & \multicolumn{2}{c|}{\textbf{2$^{\tiny\textrm{nd}}$ \layer{GRU}}} & \textbf{ICDAR-HSC2013} (\textbf{DR}/\textbf{RA}/\textbf{FM}) \\
\hline\hline
Vertical &$\downarrow$       & Horizontal &$\rightarrow$     & 0.310\,\,/\,\,0.302\,\,/\,\,0.306 \\

Vertical &$\updownarrow$     & Horizontal &$\rightarrow$     
& 0.473\,\,/\,\,0.451\,\,/\,\,0.462 \\
Vertical &$\downarrow$       & Horizontal
&$\leftrightarrow$ 
& 0.431\,\,/\,\,0.403\,\,/\,\,0.417 \\
Vertical &$\updownarrow$     & Horizontal &$\leftrightarrow$ 
& 0.801\,\,/\,\,0.786\,\,/\,\,0.793 \\
\hline
Horizontal &$\rightarrow$      & Vertical &$\downarrow$   
& 0.321\,\,/\,\,0.310\,\,/\,\,0.315 \\
Horizontal &$\rightarrow$  & Vertical &$\updownarrow$   
& 0.602\,\,/\,\,0.573\,\,/\,\,0.587 \\
Horizontal &$\leftrightarrow$      & Vertical &$\downarrow$ 
& 0.499\,\,/\,\,0.478\,\,/\,\,0.488 \\
Horizontal &$\leftrightarrow$  & Vertical &$\updownarrow$ 
& \textbf{0.827}\,\,/\,\,\textbf{0.808}\,\,/\,\,\textbf{0.817} \\
\hline\hline
\end{tabular}
\end{table}

These results are presented in Table~\ref{tab:recurrent_layer}. It is clear that the best performance is achieved when 1) exchanging horizontal line information first and then vertical, and 2) using the bi-directional setting instead of uni-directional. 


\subsection{Ablation Study on Monotone Enforcement}\label{sec:monotone}
\begin{figure}[!h]
  \centering
  \scriptsize
  \setlength{\fboxsep}{0pt}
  \setlength\tabcolsep{3pt}
  \begin{tabular}{ccc}
       \fbox{\includegraphics[width=0.4\linewidth]{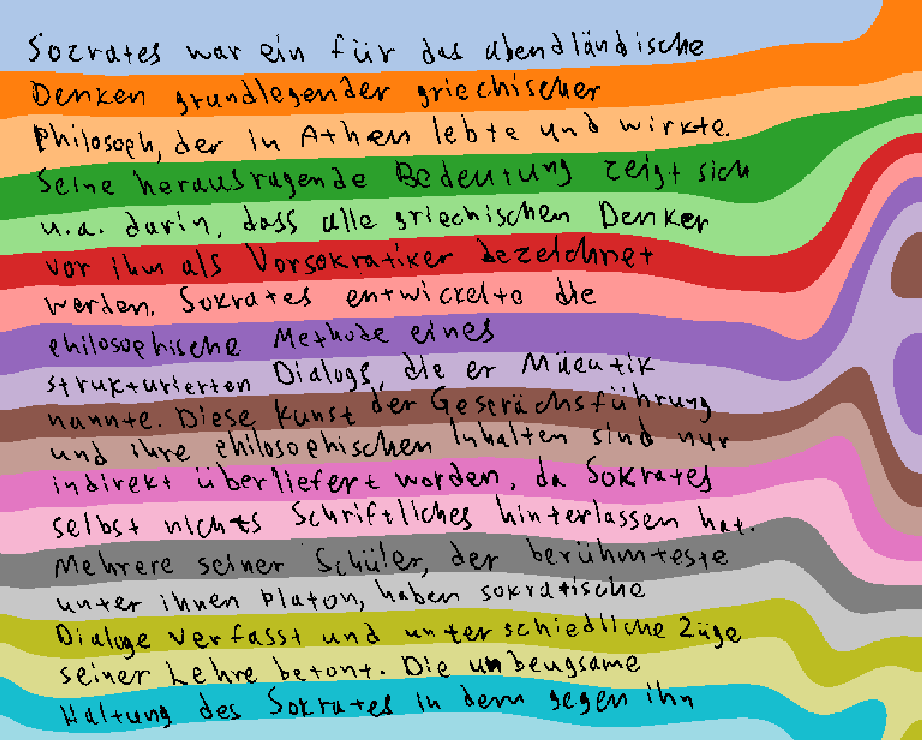}}&
       \fbox{\includegraphics[width=0.4\linewidth]{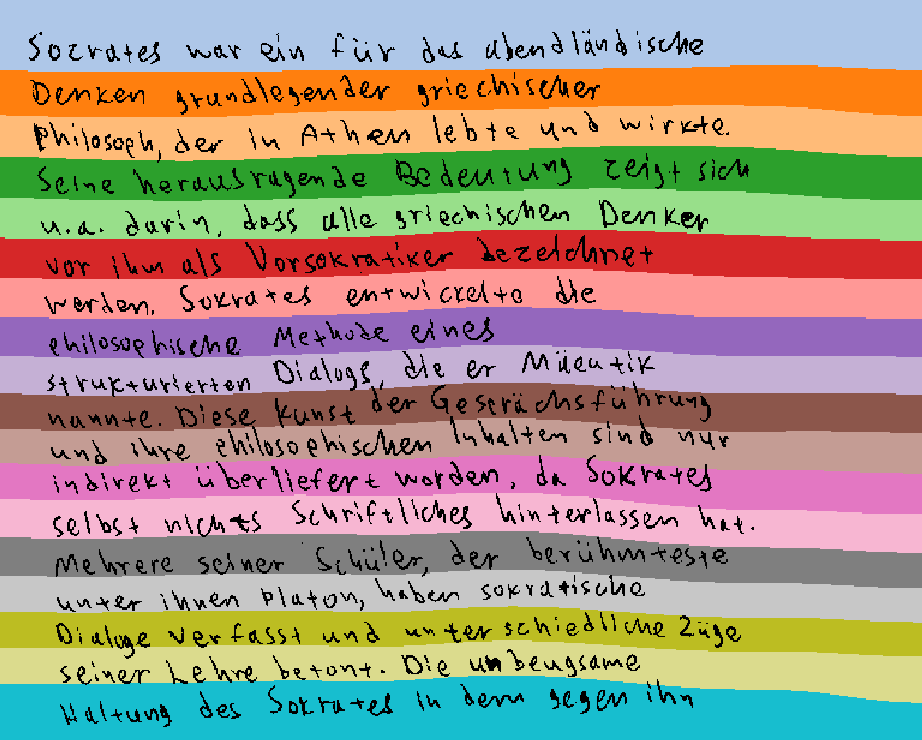}}&
       {\includegraphics[height=2.75cm]{pics/Montonic/bar.png}}\\
       (a) Baseline: w/o \texttt{cumsum}, FM=0.854 & (b) w/ \texttt{cumsum}, FM=0.973 &\\
  \end{tabular}
  \caption{Sample \layer{LineCounter} outputs w/o and w/ the \layer{cumsum} activation. 
  The output line number map is rounded and color-coded according to the rightmost coloring scheme, and texts on the map are artificially superimposed for better visualization.
  }\label{fig.cumsum}
\end{figure}

When checking the prediction from the best model of the \layer{Counter} study (referred as \textit{baseline} in this study), we notice that it fails to attain property \eqref{eq.monotonic} -- line numbers in a large background region are typically disordered (see Fig.~\ref{fig.cumsum}-(a)).

In order to attain \eqref{eq.monotonic}, we propose a simple yet effective solution by re-purposing \layer{cumsum} \eqref{eq.cumsum} as an additional activation function, but it is unclear 1) where to use it and 2) how to use it. We thus start with the baseline solution and conduct ablation studies as shown in Table~\ref{tab:Cumsum}. Two candidate locations are considered: 1) after \layer{Decoder}, \ie the last \layer{Conv2D} in \layer{Decoder}; and 2) before \layer{Decoder}, \ie the last \layer{Conv2D} in \layer{Counter}. Intuitively, it makes more sense to enforce monotone after \layer{Decoder}, but doing it before \layer{Decoder} shows better performance and helps \layer{LineCounter} attain \eqref{eq.monotonic} in practice too (see Fig.~\ref{fig.cumsum}-(b)). One possible explanation could be that the cumulative nature of \layer{cumsum} is more compatible with a recurrent module than a convolutional one. In addition, we find the best non-negative activation to be used with \layer{cumsum} is \layer{hard sigmoid}. 
Combining both leads us to the proposed \layer{LineCounter} network as shown in Fig.~\ref{fig.structure}. 

\vspace{6pt}

\begin{table}[!h]
\scriptsize
\setlength{\tabcolsep}{2pt}
\begin{center}
\caption{Ablation Study on Counting Monotone Enforcement.}\label{tab:Cumsum}
\begin{tabular}{@{}r|cc@{}}
\hline\hline
\textbf{Layers following \texttt{Conv}} & \multicolumn{2}{c}{\underline{\textbf{ICDAR-HSC2013  \,\,\,\,  (DR/RA/FM)}}} \\ 
& After \texttt{Decoder} & Before \texttt{Decoder}\\
\hline\hline
\texttt{BatchNorm}+\texttt{relu}  & \multicolumn{2}{c}{baseline: {0.827\,\,/\,\,0.808\,\,/\,\,0.817}}\\\hline
\texttt{relu}+\texttt{cumsum}& 0.364\,\,/\,\,0.183\,\,/\,\,0.244  & 0.901\,\,/\,\,0.884\,\,/\,\,0.892  \\
\texttt{abs tanh}+\texttt{cumsum} & 0.337\,\,/\,\,0.154\,\,/\,\,0.211  & 0.869\,\,/\,\,0.849\,\,/\,\,0.859  \\
\texttt{sigmoid}+\texttt{cumsum}& 0.378\,\,/\,\,0.191\,\,/\,\,0.254  & 0.931\,\,/\,\,0.914\,\,/\,\,0.922  \\
\texttt{hard sigmoid}+\texttt{cumsum}& 0.391\,\,/\,\,0.202\,\,/\,\,0.266  & \textbf{0.979}\,\,/\,\,\textbf{0.959}\,\,/\,\,\textbf{0.969}  \\
\hline\hline
\end{tabular}
\end{center}
\end{table}

\subsection{Comparisons to SoTA Solutions}
In this section, we compare the HTLS performance of the proposed \layer{LineCounter} solution with the state-of-the-art (SoTA) DNN based solutions -- the semantic segmentation based HTLS ~\cite{alberti2019labeling,barakat2018text}, the object detection based HTLS~\cite{mxnetHWR}, and the start-and-follow based HTLS~\cite{wigington2018start}. For fair comparisons, we always use the default settings of a solution provided in its public git repository, including but are not limited to pre- and post-processing, learning rate, and optimizer. 

All three public dataset introduced in Sec.~\ref{sec.ablation} are used. For each dataset and each solution, we train models with 3 random seeds and report the best performance scores in Table~\ref{tab:comparsionall}. 
As evident, object detection based HTLS~\cite{mxnetHWR} performs the worst in ICDAR-HCS2013, because many ICDAR lines are curvy or dense. All methods attain similar performance on the HIT-MW dataset, possibly because Chinese characters are more regular and the interlines in Chinese are more obvious than those in Latin and Arabic. All methods attain relatively low performance on VML-AHTE, because 1) Arabic text lines are much denser, and 2) training set is very small.


As shown in Table~\ref{tab:comparsionall}, \layer{LineCounter} outperforms the rest of approaches on all three datasets in both speed (2.5FPS) and accuracy (leading the 2nd best by 1\%-2\% in FM). More importantly, the superior performance of \layer{LineCounter} is attained by using a novel \textit{Line Counting} formulation with a relative light DNN architecture (13M parameters), instead of a bigger network or heavy post-processing. 

\begin{table}[!t]
\scriptsize
\centering
\setlength\tabcolsep{2pt}
\caption{Comparing {\scriptsize{\texttt{LineCounter}}} to SoTA HTLS DNN solutions}\label{tab:comparsionall}
\begin{tabular}{@{}c|r|r|c|c|c@{}}
\hline\hline
\multicolumn{1}{c|}{\textbf{}}&\textbf{\#Net.}&\textbf{GPU}&\multicolumn{3}{c}{\textbf{\underline{HTLS Quality Metrics (DR/RA/FM)}}} \\
\multicolumn{1}{c|}{\textbf{}}&\textbf{Param.}&\textbf{FPS}&\textbf{ICDAR-HSC2013}&\textbf{HIT-MW}&\textbf{VML-AHTE} \\

\hline
\cite{alberti2019labeling}&\textbf{12M}&1.4
& 0.955\,/\,\,0.938\,\,/\,0.946& 0.991\,/\,\,0.977\,\,/\,0.984&
0.948\,/\,\,0.928\,\,/\,0.938
\\

\cite{barakat2018text}&134M&1.9

& 0.951\,/\,\,0.940\,\,/\,0.945& 0.990\,/\,\,0.980\,\,/\,0.985&
0.932\,/\,\,0.910\,\,/\,0.921
\\

\cite{mxnetHWR}&117M&2.1

& 0.915\,/\,\,0.904\,\,/\,0.909& 0.965\,/\,\,0.951\,\,/\,0.958&
0.906\,/\,\,0.891\,\,/\,0.898
\\

\cite{wigington2018start}&15M&1.7
& 0.949\,/\,\,0.928\,\,/\,0.939& 0.988\,/\,\,0.972\,\,/\,0.980&
0.945\,/\,\,0.924\,\,/\,0.934
\\

\hline
Ours&13M&\textbf{2.5}
&\textbf{0.979}\,/\,\,\textbf{0.959}\,\,/\,\textbf{0.969}& \textbf{0.995}\,/\,\,\textbf{0.985}\,\,/\,\textbf{0.990}&
\textbf{0.958}\,/\,\,\textbf{0.938}\,\,/\,\textbf{0.948}
\\


\hline\hline
\end{tabular}
\end{table}

\section{Conclusion}
\label{ssec:subhead}

In this paper, we proposed the \textit{Line Counting} formulation for the HTLS problem. Unlike previous DNN-based problem formulations, \textit{Line Counting} is designed for the HTLS task. Consequently, it allows to directly use the HTLS line map annotation for training without extra conversions and predict the per-pixel text line map without heavy post-processing like grouping/splitting/voting. We also introduced a light yet effective end-to-end HTLS network called \layer{LineCounter} and proposed the \texttt{cumsum} activation to improve the prediction quality. Even though we have not fully optimized the architecture and hyper-parameters of \layer{LineCounter}, our initial study showed that it has already outperformed the SoTA DNN based HTLS solutions on all three public datasets (ICDAR-MSC2013, HIT-MW, and VML-AHTE). Similar formulations can also be helpful for those related tasks like word segmentation and paragraph segmentation.


\bibliographystyle{IEEEbib}

\end{document}